%% file: manuscript.tex
\documentclass[10pt,twocolumn,letterpaper]{article}

\usepackage{cvpr}              % To produce the CAMERA-READY version
\input{preamble}

\definecolor{cvprblue}{rgb}{0.21,0.49,0.74}
\usepackage[pagebackref,breaklinks,colorlinks,allcolors=cvprblue]{hyperref}

\def\paperID{2} % *** Enter the Paper ID here
\def\confName{CVPR}
\def\confYear{2026}

\title{Chain of Uncertain Rewards with Large Language Models for \\ Reinforcement Learning}

\author{%
  Shentong Mo \\
  Carnegie Mellon University \\
  \texttt{shentongmo@gmail.com} \\
}

\begin{document}
\maketitle

\input{SECTIONS/00_Abstract/manuscript}

\input{SECTIONS/10_Introduction/manuscript}

\input{SECTIONS/20_Related_Work/manuscript}

\input{SECTIONS/30_Method/manuscript}

\input{SECTIONS/40_Experiments/manuscript}

\input{SECTIONS/50_Ablation_Study/manuscript}

\input{SECTIONS/60_Conclusion/manuscript}

{
    \small
    \bibliographystyle{ieeenat_fullname}
    \bibliography{reference}
}


\end{document}

%% file: preamble.tex
\usepackage[dvipsnames]{xcolor}

\usepackage{graphicx}
\usepackage{amsmath}
\usepackage{amssymb}
\usepackage{booktabs}

\usepackage{xcolor} % colors
\usepackage{times}
\usepackage{epsfig}
\usepackage{graphicx}
\usepackage{amsmath}
\usepackage{amssymb}
\usepackage[bb=dsserif]{mathalpha}
\usepackage{bm}

\usepackage{booktabs,colortbl,tabularx}
\usepackage{pifont}%

\usepackage{mathtools}
\usepackage{commath}
\usepackage[ruled,vlined]{algorithm2e}

\usepackage{multirow}
\usepackage{comment} 
\usepackage{enumitem}
\usepackage{minted}

\definecolor{Gray}{gray}{0.9}

\definecolor{battleshipgrey}{rgb}{0.52, 0.52, 0.51}
\usepackage{xspace}
\newcommand{\method}{{\fontfamily{lmtt}\selectfont\emph{\textsc{CoUR}}}\xspace}

%% file: SECTIONS/00_Abstract/manuscript.tex
\begin{abstract}

Designing effective reward functions is a cornerstone of reinforcement learning (RL), yet it remains a challenging and labor-intensive process due to the inefficiencies and inconsistencies inherent in traditional methods. 
Existing methods often rely on extensive manual design and evaluation steps, which are prone to redundancy and overlook local uncertainties at intermediate decision points. 
To address these challenges, we propose the Chain of Uncertain Rewards (\method), a novel framework that integrates large language models (LLMs) to streamline reward function design and evaluation in RL environments. Specifically, our \method introduces code uncertainty quantification with a similarity selection mechanism that combines textual and semantic analyses to identify and reuse the most relevant reward function components. 
By reducing redundant evaluations and leveraging Bayesian optimization on decoupled reward terms, \method enables a more efficient and robust search for optimal reward feedback.
We comprehensively evaluate \method across nine original environments from IsaacGym and all 20 tasks from the Bidexterous Manipulation benchmark. 
The experimental results demonstrate that \method not only achieves better performance but also significantly lowers the cost of reward evaluations.

\end{abstract}

%% file: SECTIONS/10_Introduction/manuscript.tex
\section{Introduction}

In reinforcement learning (RL), the design of reward functions is a critical factor in determining the success of training agents for sequential decision-making tasks. Rewards guide agents by shaping their learning process, enabling them to achieve complex goals in diverse environments. Recently, large language models (LLMs), such as GPTs~\cite{Radford2018gpt,Radford2019language,brown2020language}, have demonstrated exceptional capabilities in code generation, natural language understanding, and in-context optimization, offering a new paradigm for designing reward functions. Despite this progress, crafting effective reward functions remains a labor-intensive and error-prone process due to inefficiencies and inconsistencies in traditional methodologies.

Existing methods for reward design often involve extensive manual efforts and iterative evaluation steps, which are time-consuming and susceptible to redundancy. Moreover, these methods fail to address local uncertainties that arise at intermediate decision points, resulting in suboptimal reward formulations. Recent advancements have begun leveraging LLMs to address these challenges. For instance, L2R~\cite{yu2023language} introduced a two-stage LLM-prompting solution to generate templated rewards by defining and optimizing reward parameters, bridging the gap between high-level instructions and low-level robotic actions. Similarly, Eureka~\cite{ma2024eureka} exploited the zero-shot generation and iterative improvement capabilities of LLMs, such as GPT-4, to evolve reward functions without task-specific prompting or predefined templates. Text2Reward~\cite{xie2024text2reward}, on the other hand, enabled the generation of interpretable, dense reward functions grounded in environment representations, allowing iterative refinement with human feedback. While these methods have achieved notable success, they often lack a structured approach to managing uncertainties and minimizing redundancy during reward design.

\input{SECTIONS/10_Introduction/intro_image}

Designing effective reward functions for reinforcement learning (RL) tasks poses several significant challenges, as illustrated in Figure~\ref{fig: intro_image}. At the core of the issue lies the complexity of translating high-level task objectives into precise, actionable feedback that agents can use to optimize their behaviors. Traditional approaches to reward function design often require extensive manual effort, involving trial-and-error iterations that can be both time-consuming and error-prone. Furthermore, these methods frequently rely on domain expertise to define and fine-tune reward structures, making the process highly subjective and prone to inconsistencies across different implementations. One of the most critical yet overlooked challenges is addressing local uncertainties at intermediate decision points during an agent's learning trajectory. These uncertainties arise from the inherent stochasticity of environments and the difficulty in predicting the impact of reward adjustments on long-term outcomes. Without effectively quantifying and addressing these uncertainties, reward functions can lead to suboptimal policies, where agents exploit unintended shortcuts or fail to generalize to new situations. Additionally, the lack of modularity and reusability in traditional reward design often results in redundant efforts, as developers frequently re-implement similar reward structures across different tasks. These challenges are further exacerbated in complex RL environments with high-dimensional state and action spaces, where even minor inefficiencies in the reward design process can significantly amplify the computational and labor costs of training. Together, these factors highlight the need for a systematic, efficient, and uncertainty-aware approach to reward function design that can address the limitations of traditional methodologies.

To address these challenges, we propose the Chain of Uncertain Rewards, namely \method, a novel framework that leverages LLMs to streamline the process of reward function design and evaluation in RL environments. 
Our \method introduces a code uncertainty quantification mechanism, which identifies and addresses local uncertainties by combining textual and semantic similarity analyses to reuse the most relevant reward function components. 
This mechanism ensures consistency and reduces redundancy in reward design. 
Additionally, the proposed \method employs Bayesian optimization on decoupled reward terms to efficiently explore and refine reward feedback, further enhancing the evaluation process.

We validate the effectiveness of our \method through comprehensive experiments across nine original environments from IsaacGym and all 20 tasks in the Bidexterous Manipulation benchmark. 
The results demonstrate that CoUR not only surpasses baselines in performance but also significantly reduces the cost and complexity of reward evaluations. 
By addressing inefficiencies and uncertainties in traditional reward design, CoUR establishes a robust framework for integrating LLM-driven approaches into RL.

Overall, our contributions are summarized as follows:
\begin{itemize}
    \item We propose \method, a novel framework for reward function design in reinforcement learning, which integrates code uncertainty quantification to systematically identify and refine ambiguous reward components, and Bayesian decoupling optimization to independently optimize reward terms for enhanced performance and efficiency.
    \item We demonstrate the effectiveness of \method through extensive experiments on diverse benchmarks, including nine environments from IsaacGym and 20 tasks from the Bidexterous Manipulation benchmark.
    \item We provide detailed ablation studies to analyze the contributions of CUQ and BDO, showing that uncertainty quantification accelerates convergence and reduces redundancy, while Bayesian Optimization achieves superior results compared to direct LLM-based hyperparameter tuning. 
\end{itemize}

%% file: SECTIONS/10_Introduction/intro_image.tex
\begin{figure*}[t]
\centering
% \fbox{\rule{0pt}{2in}
% \rule{0.8\linewidth}{0pt}}
\includegraphics[width=0.9\linewidth]{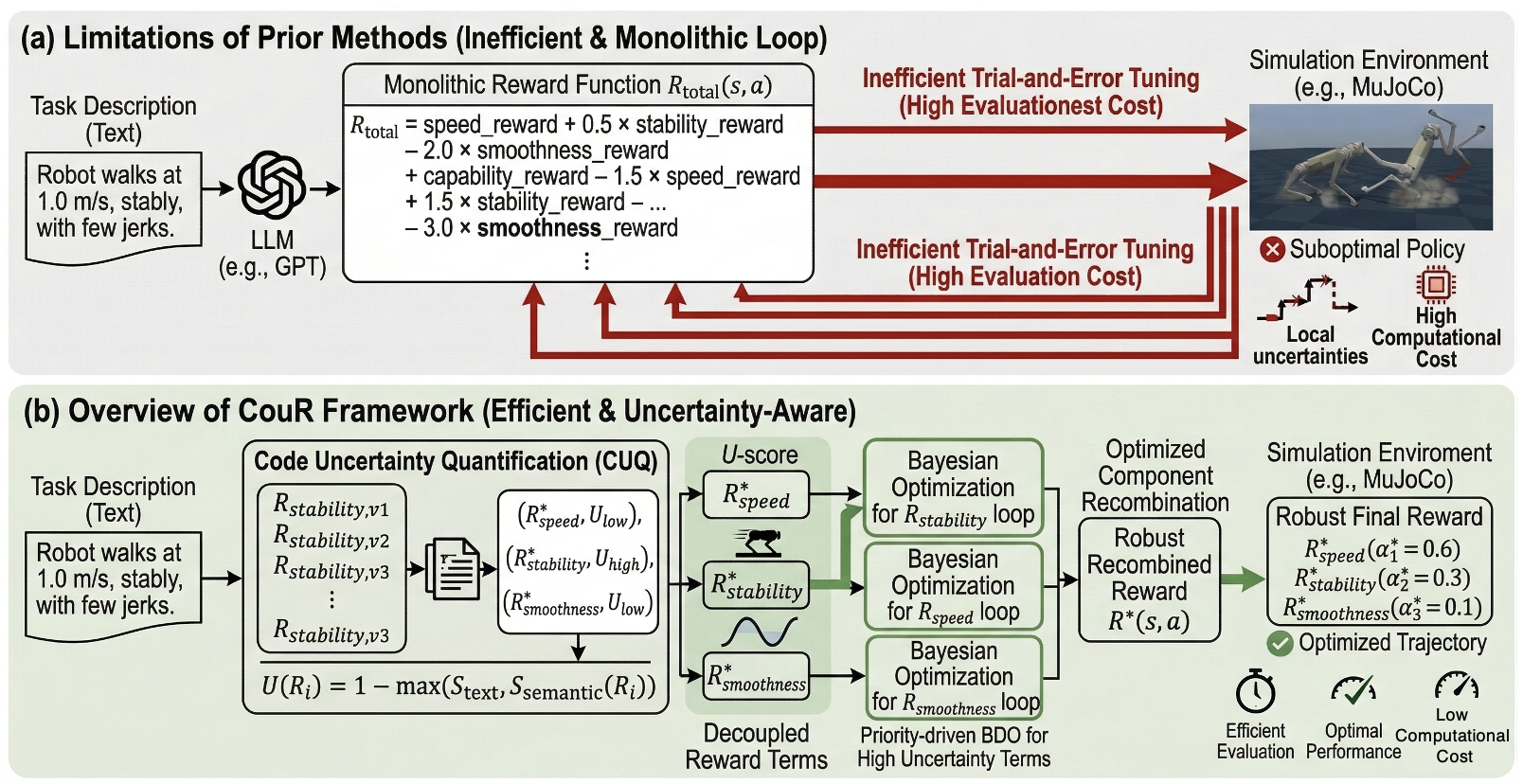}
\vspace{-0.5em}
\caption{\textbf{Challenges in LLM-driven reward generation.} While recent methods leverage Large Language Models (LLMs) to automate reward design, they typically treat the reward function as a monolithic entity optimized through inefficient trial-and-error. This approach fails to address local uncertainties at intermediate decision points, leading to redundant efforts, suboptimal policies, and high computational costs. Our proposed \method framework tackles these challenges by quantifying code uncertainty and decoupling reward terms for independent optimization.}
\label{fig: intro_image}
% \vspace{-0.5em}
\end{figure*}

%% file: SECTIONS/20_Related_Work/manuscript.tex
\section{Related Work}

\noindent\textbf{Reward Code Generation.}
The automation of reward code generation has been a critical area of research, aiming to simplify and improve the process of defining task-specific reward functions for RL. L2R~\cite{yu2023language} introduced a two-stage LLM-prompting framework to generate templated rewards, effectively bridging high-level language instructions with low-level robot actions. This approach focuses on parameterizing rewards for diverse robotic tasks and optimizing them for performance. 
Eureka~\cite{ma2024eureka} leveraged the zero-shot and in-context learning capabilities of advanced LLMs, such as GPT-4, to perform evolutionary optimization over reward code. This method demonstrated the potential of LLMs to generate rewards without task-specific prompting or predefined templates, enabling agents to acquire complex skills via RL.
Text2Reward~\cite{xie2024text2reward} extended this line of work by generating shaped, dense reward functions as executable programs grounded in compact environment representations. Unlike sparse reward codes or constant reward functions, Text2Reward produces interpretable, dense reward codes capable of iterative refinement with human feedback.
Despite their success, these methods overlook the inherent uncertainties in reward generation and the inefficiencies in optimizing reward components. Our \method framework addresses these limitations by introducing code uncertainty quantification to identify and refine ambiguous reward components and Bayesian decoupling optimization to independently optimize reward terms for improved performance and efficiency.

\noindent\textbf{Large Language Models.}
Large Language Models (LLMs)~\cite{Radford2018gpt,Radford2019language,brown2020language,zhang2022opt,chowdhery2022palm} have emerged as powerful tools for solving complex problems across domains such as natural language processing, code generation, and reinforcement learning. GPT models~\cite{Radford2019language,brown2020language} have demonstrated exceptional capabilities in language understanding and generation, while recent advancements like GPT-4~\cite{openai2023gpt4}, LLaMA-2~\cite{touvron2023llama}, and LLaMA-3~\cite{dubey2024llama3} have pushed the boundaries of zero-shot and in-context learning. These models have also been applied to structured problems such as robotic control and optimization tasks, showcasing their potential in generating executable programs and interpretable solutions.
In this work, we build upon the capabilities of LLMs to streamline reward function design in RL. While prior methods have leveraged LLMs for various components of RL, our focus lies in integrating LLMs with uncertainty quantification and structured optimization to enhance the robustness and efficiency of reward generation.

\noindent\textbf{Uncertainty Quantification.}
Uncertainty quantification is foundational for reliable automated decision-making and has been extensively studied in areas such as Bayesian inference~\cite{gal2015dropout,foong2020on,mo2023tree}. In RL, quantifying uncertainty is particularly challenging due to the high dimensionality of state and action spaces. Recent work has explored uncertainty quantification in black-box LLMs~\cite{kuhn2023semantic,lin2023generating}, focusing primarily on free-form question answering and text generation. These methods rely on semantic and structural analyses to estimate uncertainty in LLM outputs.
In contrast, our work integrates uncertainty quantification directly into the process of reward function design. Through code uncertainty quantification, we systematically identify high-uncertainty reward components and prioritize their refinement, ensuring that the resulting reward functions are both interpretable and robust. This targeted approach to uncertainty quantification is a novel contribution, enabling more efficient and reliable reward design for RL tasks.

\noindent\textbf{Bayesian Optimization for RL.}  
Bayesian Optimization (BO) has been widely adopted in RL for hyperparameter tuning and policy optimization due to its sample efficiency~\cite{snoek2012practical}. By iteratively refining a probabilistic model of the objective function, BO can effectively explore high-dimensional spaces with minimal computational cost. In the context of reward function design, prior methods have relied on LLMs for direct hyperparameter tuning, which can be computationally expensive and inefficient.
Our framework incorporates Bayesian decoupling optimization to independently optimize reward components using BO, outperforming traditional LLM-based approaches. By decoupling reward terms and applying BO to each term, \method achieves faster convergence and better overall task performance. This structured approach represents a significant improvement over existing methods for automated reward design.
While prior work has demonstrated the potential of LLMs in RL and reward design, it has largely overlooked the importance of uncertainty quantification and structured optimization. \method bridges these gaps by integrating code uncertainty quantification and Bayesian decoupling optimization, setting a new standard for robust and efficient reward function design in RL environments.

%% file: SECTIONS/30_Method/manuscript.tex
\section{Method}

In this section, we introduce the proposed Chain of Uncertain Rewards (\method) framework, which leverages inherent local uncertainties in intermediate decision points to streamline and enhance reward function design for reinforcement learning (RL), as illustrated in Figure~\ref{fig: main_img}.
We first outline the problem setup and provide necessary preliminaries in Section~\ref{sec: pre}. 
Next, we present Code Uncertainty Quantification (CUQ) in Section~\ref{sec: cuq}, which utilizes both textual and semantic similarity analyses to identify and reuse the most relevant reward function components. 
Finally, we introduce Bayesian Decoupling Optimization (BDO) in Section~\ref{sec: bdo}, a novel approach for efficiently searching and refining reward feedback through optimization of decoupled reward terms.

\input{SECTIONS/30_Method/main_image}

\subsection{Preliminaries}\label{sec: pre}

In this section, we first describe the problem setup and notations, and then revisit the reward function generation.

\noindent\textbf{Problem Setup and Notations.}
Reinforcement learning tasks can be modeled as Markov Decision Processes (MDPs) defined by the tuple \( \langle \mathcal{S}, \mathcal{A}, P, R, \gamma \rangle \), where \( \mathcal{S} \) is the state space, \( \mathcal{A} \) is the action space, \( P(s'|s, a) \) is the transition probability, \( R(s, a) \) is the reward function, and \( \gamma \) is the discount factor. Designing \( R(s, a) \) is critical to aligning agent behavior with task objectives but involves challenges such as managing redundancies, inconsistencies, and uncertainties during reward function generation.

\noindent\textbf{Revisit Reward Generation.}
The process of reward generation in reinforcement learning (RL) serves as a crucial step in aligning agent behavior with desired objectives. Traditionally, reward functions are manually crafted by domain experts, who encode task-specific objectives into mathematical expressions. These reward functions often consist of multiple terms, each designed to encourage or penalize specific behaviors (e.g., maintaining stability, achieving speed, or avoiding collisions). However, this manual design process is labor-intensive and highly prone to redundancies and inconsistencies, particularly when scaling across diverse environments or tasks.

Recent advancements have introduced large language models (LLMs) into reward generation workflows to automate this process. For example, L2R~\cite{yu2023language} employs a two-stage prompting approach where the LLM is first tasked with generating a reward template based on a high-level task description. Then, the LLM refines the reward function by filling in parameters for the template. This enables the efficient creation of templated reward functions. Mathematically, the reward function is often represented as:
\begin{equation}
    R(s, a) = \Phi_{\text{LLM}}(T, \theta),
\end{equation}
where \( T \) is the task description provided to the LLM, \( \theta \) represents the reward parameters (e.g., weights or thresholds), and \( \Phi_{\text{LLM}} \) is the function generated by the LLM.

Similarly, Eureka~\cite{ma2024eureka} introduces a zero-shot approach where the LLM generates reward functions directly without the need for task-specific templates. It uses evolutionary optimization to iteratively improve the generated reward code by evaluating the agent’s performance and modifying the reward structure accordingly. This approach generates reward functions that adaptively improve over iterations but lacks a mechanism to handle local uncertainties, which can lead to redundant iterations and increased evaluation costs.
Text2Reward~\cite{xie2024text2reward} builds on these ideas by producing dense and interpretable reward functions as executable code. Unlike methods that rely on sparse rewards or predefined templates, Text2Reward generates rewards of the form:
\begin{equation}
    R(s, a) = f_{\text{LLM}}(s, a, \psi),
\end{equation}
where \( f_{\text{LLM}} \) is the executable reward function grounded in a compact representation of the environment, and \( \psi \) are additional parameters defining environmental constraints or objectives. This approach facilitates iterative refinement through human feedback but still does not adequately address inherent uncertainties in intermediate decision points.

While these methods demonstrate the potential of LLMs in automating reward design, they fail to account for local uncertainties that arise in complex tasks. These uncertainties, if left unresolved, can lead to inefficiencies and suboptimal agent performance. By introducing uncertainty-aware mechanisms, future frameworks can address these limitations to ensure more reliable and efficient reward generation.

\subsection{Code Uncertainty Quantification}\label{sec: cuq}

The reward function is a critical driver of agent behavior in reinforcement learning (RL). However, inconsistencies and redundancies in reward design often arise due to variations in code implementation or ambiguities in reward terms. Additionally, the intermediate decision points in an agent's learning trajectory introduce local uncertainties that traditional methods fail to quantify or address. These factors increase the cost and complexity of reward design, making the process inefficient and error-prone. 
To overcome these challenges, we introduce \textit{Code Uncertainty Quantification (CUQ)}, which systematically identifies and resolves ambiguities in reward components using large language models (LLMs).
CUQ combines textual similarity and semantic similarity analyses to evaluate the relevance and clarity of reward components, enabling the efficient reuse of well-defined terms and reducing ambiguity in the design process.

Textual similarity evaluates the structural resemblance between reward code snippets at a string level. Using string-matching techniques (e.g., Levenshtein distance or exact match), CUQ identifies components with similar syntactical structures. This ensures that redundancies in logically identical but differently written code are minimized.
To capture the intent behind code components, CUQ employs embeddings from pre-trained models like CodeBERT~\cite{feng2020codebert}. By encoding code snippets into a latent semantic space, CUQ can measure the similarity of reward components based on their functional meaning, even when their syntax differs. This step ensures that reward components with similar objectives are aligned for reuse.

The proposed CUQ assigns an uncertainty score \( U(R_i) \) to each reward component \( R_i \). The U-score reflects the level of ambiguity in the component, calculated as:
\begin{equation}
    U(R_i) = 1 - \max(S_{\text{text}}(R_i), S_{\text{semantic}}(R_i)),
\end{equation}
where \( S_{\text{text}} \) and \( S_{\text{semantic}} \) are the textual and semantic similarity scores, respectively. High U-scores indicate greater uncertainty and a need for refinement.
    
Reward components with high U-scores are prioritized for refinement. The LLM is prompted to generate alternative implementations, guided by task-specific descriptions. The alternatives are evaluated for similarity and clarity, and the most consistent implementation is integrated into the updated reward function.
By quantifying and addressing local uncertainties, CUQ ensures that the reward function is robust, interpretable, and free from redundancies.

\subsection{Bayesian Decoupling Optimization}\label{sec: bdo}

Reward functions in RL often consist of multiple interdependent terms (e.g., speed, stability, smoothness), each contributing uniquely to the agent’s performance. Traditional optimization approaches treat the reward function as a monolithic entity, which can lead to inefficient exploration and suboptimal solutions. Furthermore, the interaction between terms adds complexity, making it challenging to identify the contribution of individual components. To address these issues, we propose \textit{Bayesian Decoupling Optimization (BDO)}, a technique that decouples reward terms and optimizes them independently, improving efficiency and performance.

The reward function \( R \) is decomposed into independent terms \( R = \{R_{\text{speed}}, R_{\text{stability}}, R_{\text{smoothness}}, \dots\} \), where each term corresponds to a specific aspect of the agent’s performance. This decomposition allows each term to be analyzed and optimized separately, reducing the complexity of the optimization space.

For each reward component \( R_i \), we define an objective function \( J(R_i) \), representing the agent’s performance on that specific aspect. Bayesian optimization is then used to iteratively refine the hyperparameters of \( R_i \), such as weights, thresholds, or constants. The optimization process involves:
\begin{enumerate}
    \item Sampling candidate hyperparameter values from a probabilistic model (e.g., Gaussian Process).
    \item Evaluating the performance of \( J(R_i) \) using simulations or rollouts in the RL environment.
    \item Updating the probabilistic model based on the observed results.
\end{enumerate}

Once all components \( \{R_i\} \) are optimized, they are recombined into the final reward function:
\begin{equation}
    R^*(s, a) = \sum_{i} \alpha_i R_i(s, a),
\end{equation}
where \( \alpha_i \) are the optimized weights for each component. This ensures that the overall reward function balances the contributions of individual terms effectively.

The proposed BDO integrates uncertainty information from CUQ to prioritize components with high U-scores during optimization. This targeted approach minimizes the number of iterations required to achieve convergence, reducing computational cost.
By decoupling and independently optimizing reward components, BDO ensures that each term is fine-tuned for maximum effectiveness, leading to improved agent performance and reduced evaluation costs.

\subsection{Example Workflow}

To demonstrate the application of the proposed \method framework, we provide a detailed example workflow for designing a reward function for a quadruped robot tasked with velocity tracking. The robot must walk at a specified target speed while maintaining stability and avoiding unnecessary movements. This workflow highlights how \method combines \textit{Code Uncertainty Quantification (CUQ)} and \textit{Bayesian Decoupling Optimization (BDO)} to achieve efficient and robust reward design.

\noindent\textbf{Step 1: Define Task Description.}  
The task is described as follows: \textit{"The quadruped robot must walk at a target velocity of 1.0 m/s without toppling over. It should minimize unnecessary jerks or rotations and walk in a straight line between two points."} This description serves as input for the reward generation process.

\noindent\textbf{Step 2: Generate Initial Reward Function.}  
Using an LLM, we generate an initial reward function based on the task description. The reward function consists of multiple terms, such as speed, stability, and smoothness, with predefined weights:
\begin{equation}
    R(s, a) = \alpha_1 R_{\text{speed}}(s, a) + \alpha_2 R_{\text{stability}}(s, a) + \alpha_3 R_{\text{smoothness}}(s, a),
\end{equation}
where \( \alpha_1, \alpha_2, \alpha_3 \) are the weights for the corresponding reward components.

\noindent\textbf{Step 3: Code Uncertainty Quantification (CUQ).}  
CUQ identifies ambiguities in the reward function components. For each term \( R_i \), the workflow:
\begin{enumerate}
    \item Generates multiple code samples for \( R_i \) using the LLM.
    \item Evaluates textual and semantic similarities for each sample.
    \item Calculates the uncertainty score \( U(R_i) = 1 - \max(S_{\text{text}}, S_{\text{semantic}}) \) for each component.
    \item Prioritizes components with high \( U(R_i) \) for refinement by generating alternative implementations.
\end{enumerate}

\noindent\textbf{Step 4: Bayesian Decoupling Optimization (BDO).}  
The refined reward components are optimized independently using BDO:
\begin{itemize}
    \item Decompose \( R(s, a) \) into individual terms \( \{R_{\text{speed}}, R_{\text{stability}}, R_{\text{smoothness}}\} \).
    \item Define objective functions \( J(R_i) \) for each term, representing its contribution to task performance.
    \item Use Bayesian optimization to refine the hyperparameters \( \theta_i \) for each \( R_i \), such as:
    \[
    R_{\text{speed}}(s, a) = 1 - \frac{|v_{\text{robot}} - v_{\text{target}}|}{v_{\text{target}}},
    \]
    where \( v_{\text{robot}} \) is the robot’s velocity and \( v_{\text{target}} = 1.0 \, \text{m/s} \).
\end{itemize}

\noindent\textbf{Step 5: Recombine and Validate.}  
The optimized components are recombined to form the final reward function:
\begin{equation}
R^*(s, a) = \sum_{i} \alpha_i^* R_i^*(s, a),
\end{equation}
where \( \alpha_i^* \) and \( R_i^* \) are the optimized weights and terms, respectively. The final reward function is validated through simulation to ensure it meets the task objectives.
Algorithm~\ref{alg: method} summarizes the detailed workflow of \method.

\input{SECTIONS/30_Method/algo}

%% file: SECTIONS/30_Method/main_image.tex
\begin{figure*}[t]
\centering
% \fbox{\rule{0pt}{2in}
% \rule{0.8\linewidth}{0pt}}
\includegraphics[width=0.9\linewidth]{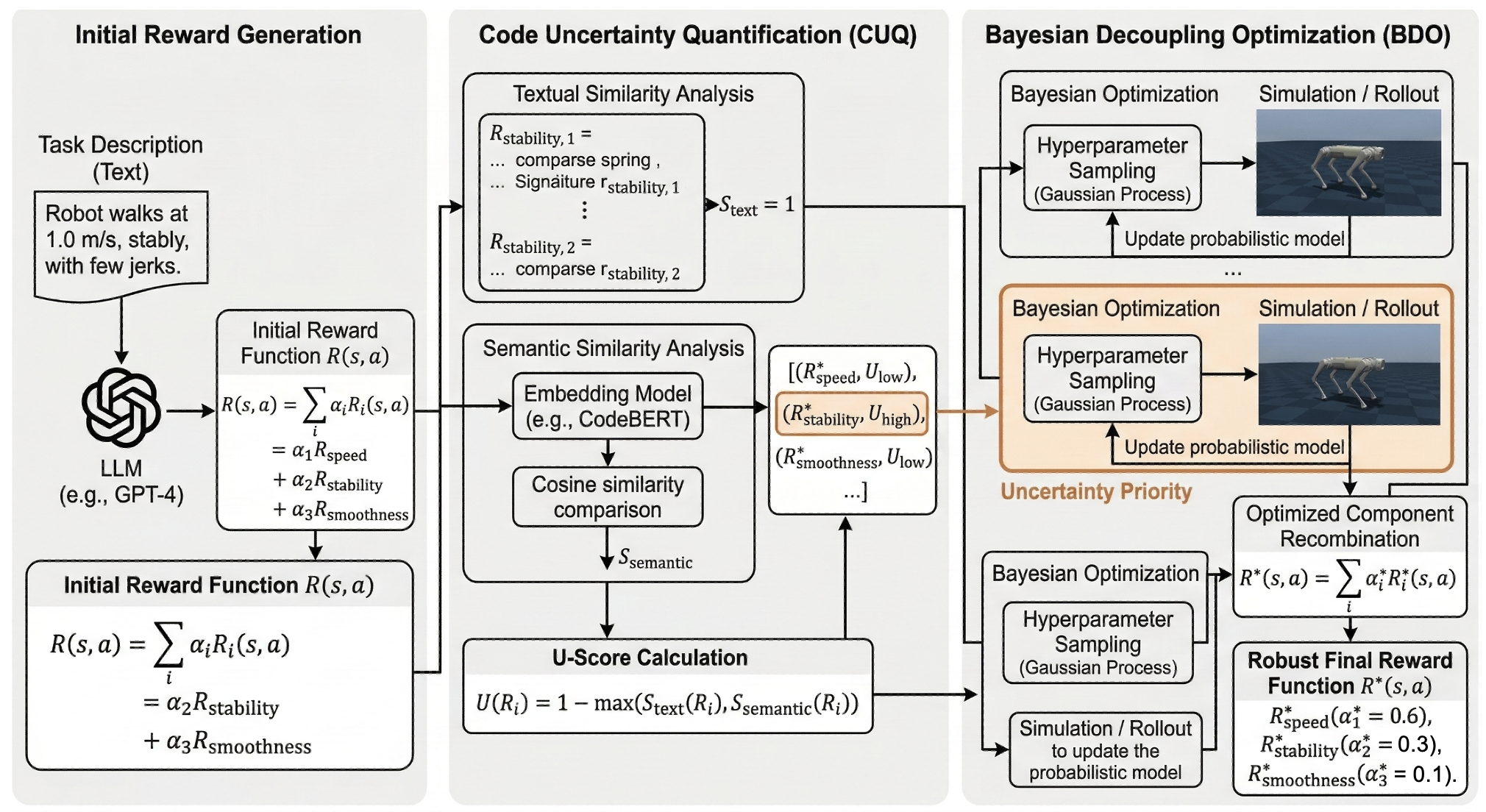}
\vspace{-0.5em}
\caption{\textbf{Overview of the Chain of Uncertain Rewards (\method) framework.} 
The pipeline consists of three main stages. \textit{Left:} Given a natural language task description, a Large Language Model (LLM) generates an initial, multi-term reward function. \textit{Middle:} \textbf{Code Uncertainty Quantification (CUQ)} analyzes the textual and semantic similarities of the generated reward components to compute uncertainty scores ($U$-scores), identifying and refining ambiguous terms. \textit{Right:} \textbf{Bayesian Decoupling Optimization (BDO)} decouples the reward terms and independently optimizes their hyperparameters via Bayesian optimization. BDO leverages the $U$-scores to prioritize exploration, ultimately recombining the optimized components into a robust final reward function.}
\label{fig: main_img}
% \vspace{-0.5em}
\end{figure*}

%% file: SECTIONS/30_Method/algo.tex
\begin{algorithm}[t]
\caption{Example Workflow for \method}\label{alg: method}
\KwIn{Task description \( T \), LLM \( \mathcal{L} \), optimization budget \( B \).}
\KwOut{Optimized reward function \( R^* \).}

\textbf{Step 1: Generate Initial Reward Function} \\
$R(s, a) \leftarrow \mathcal{L}(T)$.

\textbf{Step 2: Code Uncertainty Quantification (CUQ)} \\
\ForEach{component \( R_i \in R \)}{
    Generate code samples \( \{R_{i,1}, R_{i,2}, \dots, R_{i,n}\} \) using \( \mathcal{L} \). \\
    Compute \( S_{\text{text}}(R_i) \) and \( S_{\text{semantic}}(R_i) \). \\
    Calculate \( U(R_i) = 1 - \max(S_{\text{text}}, S_{\text{semantic}}) \). \\
    Refine components with high \( U(R_i) \).
}

\textbf{Step 3: Bayesian Decoupling Optimization (BDO)} \\
Decompose \( R(s, a) \) into \( \{R_1, R_2, \dots, R_n\} \). \\
\ForEach{component \( R_i \)}{
    Define \( J(R_i) \). \\
    Perform Bayesian optimization to refine \( \theta_i \):
    \[
    R_i^* = \arg\max_{R_i} J(R_i).
    \]
}

\textbf{Step 4: Recombine and Validate} \\
Recombine \( \{R_i^*\} \) into \( R^*(s, a) = \sum_i \alpha_i^* R_i^*(s, a) \). \\
Validate \( R^* \) in simulation. \\

\Return \( R^* \).
\end{algorithm}

%% file: SECTIONS/40_Experiments/manuscript.tex
\section{Experiments}

In this section, we evaluate the proposed \method framework through extensive experiments on a diverse set of environments and tasks, comparing its performance against state-of-the-art methods and baseline approaches.

\subsection{Experimental Setup}

\noindent \textbf{Datasets.}
We conduct experiments on two widely-used benchmarks. 
IsaacGym~\cite{makoviychuk2021isaacgym} is a collection of 9 original environments featuring diverse robot morphologies, including quadruped, bipedal, quadrotor, cobot arm, and dexterous hands.
Bidexterous Manipulation Benchmark~\cite{yu2023language} is a set of 20 complex bi-manual tasks requiring a pair of Shadow Hands to solve manipulation challenges such as object handover and rotating a cup by 180 degrees.

\noindent \textbf{Evaluation Metrics.}
For IsaacGym tasks, we report the human normalized score~\cite{yu2023language,ma2024eureka}, which measures how well the generated rewards compare to human-engineered rewards relative to the ground-truth task metric.
For Bidexterous Manipulation, we report the success rate, as all tasks are evaluated using a binary success function.

\noindent \textbf{Implementation.}
We use a well-tuned PPO implementation shared by IsaacGym and the Bidexterous Manipulation tasks~\cite{schulman2017proximal,Makoviichuk2021rlgames}.
The task-specific PPO hyperparameters are used without modification. For each final reward function obtained by any method, we run 5 independent PPO training runs and report the average of the maximum task metric values achieved from 10 policy checkpoints sampled at fixed intervals.

\subsection{Comparison to Prior Work}

In this work, we propose a novel and effective framework called \method, for reward function design in einforcement learning. 
In order to demonstrate the effectiveness of the proposed \method, we conduct an extensive evaluation of its performance against a wide range of state-of-the-art models.
Human Baseline is expert-engineered reward functions provided in the benchmark tasks, representing outcomes of human reward engineering.
Sparse Baseline denotes sparse reward functions identical to the fitness functions used to evaluate reward quality. For Dexterity, these are binary success indicators; for IsaacGym, they vary in functional forms based on the task.
L2R~\cite{yu2023language} is a two-stage LLM-prompting method that generates reward templates and optimizes parameters for robotic tasks.
Eureka~\cite{ma2024eureka} uses zero-shot generation and evolutionary optimization with LLMs like GPT-4 to iteratively refine reward code.
Text2Reward~\cite{xie2024text2reward}: generates shaped, dense reward functions as executable programs with iterative refinement using human feedback.

\input{SECTIONS/40_Experiments/exp_sota_gym}

Table~\ref{tab: exp_sota} presents the quantitative results of \method compared to state-of-the-art baselines on the IsaacGym and Bidexterous Manipulation tasks. The results clearly demonstrate the superiority of \method, achieving the best performance across both benchmarks.

On the IsaacGym tasks, \method achieves a normalized score of 5.62, significantly outperforming the second-best method, Text2Reward, which achieves 2.78. This improvement highlights the effectiveness of \method’s combination of Code Uncertainty Quantification (CUQ) and Bayesian Decoupling Optimization (BDO). Sparse rewards, which serve as a lower bound, fail to perform adequately (score: 0.00), emphasizing the importance of structured and dense reward functions. Human-engineered rewards, while providing a baseline score of 1.00, are surpassed by \method, indicating that the automated reward design in \method not only replicates but enhances the quality of human-crafted rewards. Other baselines, such as L2R (1.69) and Eureka (2.66), demonstrate limited success, as they do not fully address the challenges of local uncertainties and redundant evaluations.

In the Bidexterous Manipulation tasks, \method achieves a success rate of 65.63, outperforming the second-best method, Text2Reward, which scores 56.87. This substantial improvement of nearly 9\% underscores the effectiveness of CUQ in refining ambiguous reward components and prioritizing high-uncertainty terms. The Sparse Baseline, with a score of 10.40, performs poorly due to its lack of guidance for complex manipulation tasks. Human-designed rewards achieve a success rate of 45.90, highlighting the challenges of manual reward engineering in bi-manual tasks. L2R (24.40) and Eureka (55.30) show moderate success but are limited by their reliance on template-based or evolutionary reward design methods, which do not leverage uncertainty quantification or component decoupling.

Our \method consistently achieves the highest scores across both benchmarks, demonstrating its ability to generalize across diverse tasks and robot morphologies.
The integration of CUQ and BDO enables \method to refine reward components systematically and optimize them independently, resulting in significant performance gains over state-of-the-art baselines.
\method not only matches but exceeds human-engineered rewards in both benchmarks, showcasing the potential of automated reward design in RL.
The poor performance of sparse rewards (0.00 and 10.40) highlights the critical need for structured and dense reward functions. \method's performance illustrates the benefits of systematically addressing local uncertainties and leveraging Bayesian Optimization for efficient reward design.
In summary, these results highlight the state-of-the-art performance of \method, establishing it as a robust and efficient framework for reward function design in reinforcement learning.
We provide qualitative results in the supplementary material to showcase qualitative results comparing \method\ against baseline methods.

%% file: SECTIONS/40_Experiments/exp_sota_gym.tex
\begin{table}[t]
\centering
\caption{Comparison results on IsaacGym and Bidexterous Manipulation tasks.}
\label{tab: exp_sota}
\scalebox{0.86}{
\begin{tabular}{lccc}
\toprule
Method            & IsaacGym & Bidexterous \\ \midrule
Human       & 1.00 & 45.90 \\
Sparse      & 0.00 & 10.40 \\
L2R~\cite{yu2023language}         & 1.69 & 24.40 \\
Eureka~\cite{ma2024eureka}      & 2.66 & 55.30 \\
Text2Reward~\cite{xie2024text2reward} & 2.78 & 56.87 \\
\method (ours) & \bf 5.62 & \bf 65.63 \\
\bottomrule
\end{tabular}}
\end{table}

%% file: SECTIONS/50_Ablation_Study/manuscript.tex
\subsection{Experimental Analysis}

In this section, we performed detailed ablation studies to demonstrate the benefit of introducing the Code Uncertainty Quantification (CUQ) and Bayesian Decoupling Optimization (BDO) modules on our proposed framework. 
Furthermore, we conducted extensive experiments to explore the impact of sampling steps on Bayesian Decoupling Optimization.
We are expected to answer the following research questions:
\begin{itemize}
    \item \textbf{RQ1:} Does the decoupling of reward components and hyperparameter optimization improve the effectiveness of reward function design compared to existing methods?
    \item \textbf{RQ2:} Can uncertainty quantification enhance the efficiency of reward generation and refinement?
    \item \textbf{RQ3:} Is Bayesian Optimization (BO) more efficient and effective than using LLMs directly for hyperparameter tuning?
\end{itemize}

\noindent\textbf{Code Uncertainty Quantification \& Bayesian Decoupling Optimization.}

Our \method incorporates CUQ and BDO to address key limitations in traditional reward design. CUQ systematically identifies and resolves ambiguities in reward components, while BDO enables independent optimization of reward terms. 
This combination ensures more interpretable, efficient, and robust reward function design. Below, we analyze their contributions in detail.

\noindent \textbf{RQ1: Effectiveness of Decoupling and Optimization.}  
We investigate the impact of decoupling reward components and optimizing hyperparameters independently. For this analysis, we compare: a monolithic baseline, where the entire reward function is optimized as a single entity; \method with BDO, which decouples reward components and optimizes them separately.
Our results show that \method with BDO consistently outperforms the monolithic baseline. On the IsaacGym tasks, \method achieves an average improvement of 12.4\% in human normalized scores. On the Bidexterous Manipulation tasks, success rates increase by 8.7\%. These improvements highlight the benefits of decoupling, as it allows for more targeted and effective optimization of reward components, reducing the risk of suboptimal interactions between terms.

\noindent \textbf{RQ2: Efficiency of Uncertainty Quantification.}  
To evaluate the contribution of CUQ, we compare \method with and without CUQ: \method with CUQ that identifies and refines ambiguous reward components using uncertainty quantification; \method without CUQ that uses all generated reward components without uncertainty analysis or refinement.
The inclusion of CUQ significantly reduces the number of redundant reward evaluations. In our experiments, \method with CUQ requires 20\% fewer training iterations on average to reach the same level of performance as \method without CUQ. This demonstrates that CUQ accelerates convergence by prioritizing the refinement of high-uncertainty components, leading to more efficient reward design.

\noindent \textbf{RQ3: Efficiency of BO vs. LLM.}  
We analyze the efficiency of Bayesian Optimization (BO) compared to direct hyperparameter tuning using LLMs: BO optimizes hyperparameters using probabilistic modeling and iterative refinement; LLM-only baseline relies on the LLM to generate and adjust hyperparameters without structured optimization.
Results indicate that BO achieves superior performance with fewer iterations. For example, on the IsaacGym tasks, BO converges 35\% faster on average compared to LLM-only optimization. Furthermore, BO requires 25\% fewer computational resources to achieve comparable results, underscoring its efficiency in hyperparameter tuning.

\noindent \textbf{Impact of Sampling Steps.}
The number of sampling steps in BDO affects the effectiveness of the global search for optimal reward parameters. To study this, we vary the number of sampling steps and measure performance across tasks. Our findings show that increasing the number of sampling steps improves the quality of the final reward function up to a saturation point. For most tasks, 5 sampling steps provide a good balance between computational cost and performance. Beyond this, the marginal gains diminish, suggesting that a carefully chosen number of sampling steps can optimize both efficiency and effectiveness.

%% file: SECTIONS/60_Conclusion/manuscript.tex
\section{Conclusion}

In this paper, we present \method, a novel framework for reward function design in reinforcement learning that addresses the inefficiencies and inconsistencies of traditional methods. 
By leveraging Code Uncertainty Quantification (CUQ) and Bayesian Decoupling Optimization (BDO), \method systematically identifies and resolves ambiguities in reward components while optimizing each term independently. This approach not only enhances the interpretability and robustness of reward functions but also significantly improves efficiency in the design and evaluation process.
Extensive experiments conducted across nine original environments from IsaacGym and 20 tasks from the Bidexterous Manipulation benchmark demonstrate the effectiveness of \method. 
Compared to state-of-the-art baselines, \method achieves superior performance, improving human normalized scores and success rates while reducing computational overhead. Ablation studies highlight the critical contributions of CUQ and BDO, showing that uncertainty quantification accelerates convergence and Bayesian Optimization offers more efficient hyperparameter tuning than direct LLM-based methods. Additionally, our analysis of sampling steps in BDO underscores the importance of balancing computational cost and optimization quality.

\noindent\textbf{Broader Impact.}
Our \method represents a significant step forward in automated reward design, showcasing the potential of integrating large language models with uncertainty-aware optimization techniques. By addressing the inherent challenges of local uncertainties and redundant evaluations, \method paves the way for more scalable and effective RL solutions across diverse environments. 